\RequirePackage{fix-cm}
\documentclass[smallextended]{svjour3r}       
\smartqed  
\usepackage{graphicx}
\usepackage{epstopdf}
\usepackage{mathtext}
\usepackage{amssymb,amsmath,amstext,amscd,latexsym,amsfonts,array,epsfig}
\usepackage{mathrsfs}
\usepackage[matrix,arrow,curve]{xy}
\usepackage{xypic}
\usepackage{color}
\usepackage{cmll}
\usepackage[cp1251]{inputenc}
\usepackage[T2A]{fontenc}
\usepackage{ruslist}
\begin{document}

\title{A Way to Facilitate Decision Making in a Mixed Group of Manned and Unmanned Aerial Vehicles
\thanks{The project was partly supported by RFBR grant 16-08-00832a}
}
\author{Dmitry Maximov         \and
        Yury Legovich            \and
        Vladimir Goncharenko
}


\institute{Dmitry Maximov \at
              Trapeznikov Institute of Control Science Russian Academy of Sciences, 65, Profsoyuznaya st., Moscow, Russia, 117997 \\
              Tel.: +7-495-3348721\\
              \email{dmmax@inbox.ru}           
           \and
           Yury Legovich \at
              Trapeznikov Institute of Control Science Russian Academy of Sciences, 65, Profsoyuznaya st., Moscow, Russia, 117997
              \and
           Vladimir Goncharenko \at
              Moscow Aviation Institute, 4, Volokolamskoe highway,  Moscow, Russia, 125993
}

\maketitle

\begin{abstract}
A mixed group of manned and unmanned aerial vehicles is considered as a distributed system. A lattice of tasks which may be fulfilled by the system matches to it. An external multiplication operation is defined at the lattice, which defines correspondingly linear logic operations.  Linear implication and tensor product are used to choose a system reconfiguration variant, i.e., to determine a new task executor choice. The task lattice structure (i.e., the system purpose) and the operation definitions largely define the choice. Thus, the choice is mainly the system purpose consequence. Such a method of the behavior variant choice facilitates the decision making by the pilot controlling the group. The suggested method is illustrated using an example of a mixed group control at forest fire compression.
\end{abstract}


\section{Introduction}
\label{intro}
At present, aviation surveillance systems in the emergency zone have received wide distribution \cite{gonch1}. Lately, unmanned aerial vehicles (UAV) are actively used in these surveillance systems.
Usually, single such vehicles are used that is not very efficiently, especially in large territories inspection. So, UAV group applying increases the efficiency of the task execution. However,  in this case, additional problems appear, which are connected with actions coordination.

Thus, the task of the UAV group control is of high importance. The multi-agent technology using is possible for such a UAV group control \cite{amel}. Every UAV is considered as an intelligent agent in the case.

As it is shown in \cite{gonch3}, the main item, in this case, is effective targeting for control objects and, correspondingly, the system structure reconfiguring choice. Therefore, a new approach to the goal assignment task decision is considered in the paper, correspondingly, with the system structure change choice.

There are several approaches to such a system structure control.

\textbf{The collective control} in the iterative optimization method of system element actions~\cite{kalyaev}, \cite{vv_17}. The approach demands models of system elements and an environment. The state is described then by the recursive equation system with the constraints by discrete time, element actions and the environment reaction. At each step, system elements choose actions from some admissible set in such a way, that the system state remains in the proper set. Also the quality functional increment should be extremal up to the moment when no new actions give the functional increment. The main difficulty of the approach is getting adequate efficiency estimations.

\textbf{The behavioral approach} supposes a finite situation set with which the system may have a deal. Correspondingly, there is a set of system element action models in the situations~\cite{vv_85}, \cite{vv_135}, \cite{vv_141}, \cite{vv_166}. Each element identifies the situation and chooses the behavior variant independently, but the information exchange may be used for action coordination.

\textbf{The market economy methods}~\cite{Dias}, \cite{vv_101}, \cite{vv_117}, \cite{vv_174}. In this case, an auction is held when the new task appears. All system elements appreciate their efficiency in the task decision during the auction and exchange their rate information with the others. Then the rates are revised, and the auction continues up to the moment of consensus. Then the element with the extremal rate became the executor. The same situation with several tasks targeting. In the case, the quality criterion is the maximal total reward.

\textbf{Fuzzy logic methods}~\cite{vv_81}, \cite{vv_124}, \cite{vv_134}, \cite{vv_159}. In such methods, the input variables (e.g., the variable which corresponds to a new task) transform to fuzzy linguistic variables (e.g, ``unimportant'', ``important'', ``critical'') based on membership functions use. Then, linguistic variables transform into fuzzy control solutions based on fuzzy inference rules. The latter ones then convert to concrete output variable values. The fuzzy inference rule is formulated in the following way:

$R_{i}:\;$if $x_{\mu}\in X_{mu}^{i}$, then $u_{\mu}\in U_{mu}^{i}$.

Here $x_{\mu}$ is an input linguistic variable. $X_{mu}^{i}$ is a fuzzy linguistic term (a fuzzy set of the variable $x_{\mu}$ values). $u_{\mu}$ is a fuzzy conclusion of the rule $R_{i}$ which corresponds to the term $X_{mu}^{i}$ of the input linguistic variable and to the term $U_{mu}^{i}$ of the output linguistic variable. $i$ indexes the set of fuzzy inference rules. The main difficulty here is the absence of determination methods of membership functions, inference rules, set bounds of fuzzy values, their types and so on. All these are determined based on experts opinions.

\textbf{The gregarious (swarm) approach}~\cite{kalyaev}, \cite{vv_20}, \cite{vv_60}. The main feature of the systems focused on the  gregarious approach is the absence of communication channels between system elements. Behavior monitoring of available items is the unique information source about  actions of other system elements. The system elements can coordinate their actions assessing the behavior of neighbors. The duration of task solving processes is significantly increased in such systems compared with the other types.

The methods considered in the article are close to the behavioral approach and to fuzzy logic methods. The system elements have their action variant sets as in the behavioral approach, but a communication network connects them. Also, decision making about the action variant choice is made from the many-valued implication estimation, it is not based on the fuzzy inference. The inference rules are determined here by the structure of the task set which the system can fulfill. I.e., they are the system internal property which does not depend on the expert mind. Thus, agent behavior is investigated as an emergent system property without any modeling of the agent, the environment and strategies \cite{brooks}. The system can determine its behavior from its internal structure and, correspondingly, its purpose, that is the main difference and the advantage of the proposed approach from the rest.

We use the term ``desires'' for goals which the agent would like to achieve. Also, we use the term ``intentions'' for those goals from desired which the agent can achieve. Both these terms are taken from BDI paradigm (belief-desire-intention, \cite{rao}), but the agents' ways to determine desires and intentions are out of the research bounds.

All the system states, or tasks which the system can fulfill, are represented as a lattice with the system tasks as generators (like in \cite{maximov1}, \cite{maximov2}). Thus, the generators are the group desires. The set of elementary action which are necessary to fulfill the tasks determines these tasks. Therefore, it is possible to refer to the tasks as to sets. The generator joins correspond to system functioning variants with different task sets to fulfill. Thus, the more tasks are solved in this variant, the higher the correspondent join lies in the task lattice. Some of these joins are the intentions of an agent or the whole group.

The generator meets are subtasks or actions, which are included in different tasks. The bottom lattice element (0) corresponds to inaction, and the top one corresponds to the join of all the tasks fulfilling. Overlying elements in the lattice diagram are more valuable than underlying ones since the more active system behavior is more valuable than less active one. Otherwise, the more intentions are included in the element, the higher the element value is.

The choice of a new task executor was carried out in \cite{maximov1} by the estimation of a many-valued implication. The latter was determined at the lattice based only on its structure, in the case of the lattice was Brouwer (equivalently, distributive for a finite lattice). However, in general, it is not sufficient since the task lattice in not always distributive. Also, an unambiguous implication estimate does not provide sufficient flexibility in decision making.

The element multiplication was additionally determined at an arbitrary lattice in\cite{maximov2}. That allowed to get an intelligent behavior in a group of robot-janitors by the same method of implication estimating. However, in this case, the implication is the linear one. Also, the linear implication has also been used  in \cite{ieee} to choose the variant of control system functioning during car unloading in the port.

In this paper, the element multiplication is also determined at the lattice. That allows to determine multiplicative linear logic operations in addition to lattice ones. Then, the linear implication is considered as switching from one task to another. The implication is also a lattice element and has some degree of value. Thus, we can appreciate the switching value of different tasks to some new one. Hence, we present the implication $a\multimap b = c$, with $a, b, c$ are the task lattice elements, as a transition from task \emph{a} execution to task \emph{b} execution with the value degree \emph{c}.

The lattice multiplication may be determined in different ways. In the paper, we improve  the \cite{maximov2} reasons which allow to determine the most of different element multiplications. The reasons are related to the determination of duality and requirements to open and closed fact classes. The requirements limit the products of elements not always uniquely. Hence, it is possible to assess the lattice element multiplications with more than one value that allows to choose the system transformation variant more flexible than in many-valued logic \cite{maximov1}.

Unlike the previous work \cite{maximov2}, in this paper, we consider configuration changes in the case of tasks which fulfil in parallel. It means that tensor products are included in the first and second members of the implication.

The suggested approach is illustrated by an example of an executor choosing in a mixed group of one manned and unmanned aerial vehicles when extinguishing forest fires.

\section{Mathematical Backgrounds \cite{maximov2}, \cite{ieee}}
Suppose that a multi-agent system has a list of tasks, i.e., goals its agents are able to work
on. Otherwise, the goals are called agents'desires. We call these tasks independent. The tasks are the generators of the lattice which consist of all tasks the system can fulfill. The agents also have subtasks that are not independent and can be a part of different independent tasks.

A more general notion than a task is the notion of a system task or system state. In this case, the system fulfills one or more tasks with at least one agent. The system state is determined not only by tasks fulfilling now but also by agent intentions. Otherwise, by the desires, the agents are intended (i.e., are able) to fulfill in future. The system state is the generator join in the task lattice. We will also talk about the system state as about the execution of certain processes. In this case, the process resource is considered as intentions of agents which fulfill a certain task. Not only one process may be included in the system state. It means that the system state (i.e., the generator join) may be broken into smaller groups.

The linear implication at the task lattice describes an agent transition from the execution of one task to another one. We use the notion of the complete lattice to represent the system states, i.e., the set of tasks the system can fulfill.
\newtheorem{Definition}{definition}
\begin{Definition} A \textbf{partially-ordered set} $P$ is the set with such a binary relation  $x \leqslant y$ for elements in it, that for all $x, y, z \in P$ the next relationships are performed:
\begin{itemize}
\item	$x \leqslant x$ (reflexivity);
\item	if $x \leqslant y$ and $y \leqslant x$, then $x = y$ (anti-symmetry);
\item	if $x \leqslant y$ and $y \leqslant z$, then $x \leqslant z$ (transitivity).
\end{itemize}
\end{Definition}

This means that in the partially-ordered set not all elements are comparable with each other. This property distinguishes these sets from linear-ordered ones, i.e., from numeric sets which are ordered by a norm.  Thus, the elements of the partially-ordered set are the objects of more general nature than numbers. In the partially-ordered set diagram, the greater the element (i.e., vertex, node) is the higher it lies, and the elements are comparable with each other lie in the same path from a less element to a greater one. An example of a partially-ordered set diagram is represented in Fig.~\ref{fig1} (Sec.~\ref{3}) which is also a lattice diagram.
\begin{Definition}The \textbf{upper bound} of a subset $X$ in a partially-ordered set $P$ is the element $a\in P$ which contains all $x\in X$.
\end{Definition}

The supremum or \emph{join} is the smallest subset $X$ upper  bound. The infimum or \emph{meet} defines dually as the greatest element $a\in P$ contained in all $x\in X$.
\begin{Definition} A \textbf{lattice} is a partially-ordered set, in which every two elements have their meet, denoted as $x\wedge y$, and join, denoted as $x\vee y$.
\end{Definition}

In the lattice diagram the elements join is the nearest upper element to both of them, and the meet is the nearest lower one to both. The elements which generate by joins and meets all other elements are called \textbf{generators}.

They refer to the lattice as the \emph{complete} lattice if its arbitrary subset has the join and the meet. Thus, any complete lattice has the greatest element ``$\top$'' and the smallest one ``0'' and every finite lattice is complete. Distributivity identities on join and meet operations are satisfied in \emph{distributive lattices}.

\subsection{Elements of Linear Logic}\label{2}
If a multiplication operation is additionally defined at the lattice elements, then the operations of linear logic also exist at the lattice. We use the phase semantic of linear logic from \cite{girar}.
\begin{Definition}
A \textbf{phase space} is a pare $(M,\bot)$, where $M$ is a multiplicative monoid (i.e., a triple $(M_{0}, \cdot, e)$ with $M_{0}$ is a set and $\cdot$ is a multiplication with the unit $e$), which is also a lattice, and the element \emph{false} of the lattice $\bot\subset M$ is an arbitrary subset of the monoid.
\end{Definition}

In linear logic, the element \emph{false} differs from 0 (the minimal lattice element) in general in contrast to classical logic or intuitionistic one. The multiplication $X\cdot Y = \{x\cdot y|x\in X, y\in Y\}$ is defined for arbitrary monoid subsets (i.e. the lattice elements) $X, Y \subset M$. The linear implication $X\multimap Y = \{x|x\cdot z\in Y, \forall x\in X\}$ is also defined. For $X\subset M$ its dual is defined as $X^{\bot}\multimap \bot$. The dual element is a generalization of the negation in the case of linear logic.
\begin{Definition}
\textbf{Facts} are such subsets $X\subset M$ that $X^{\bot\bot} = X$ or equivalently $X = Y^{\bot}$ for some $Y \subset M$.
\end{Definition}

Thus, facts are lattice elements coinciding with their double negations. E.g. $\bot^{\bot} = I = \{e\}^{\bot\bot};\; \top = M = \emptyset^{\bot};\; 0 = \top^{\bot} = M^{\bot} = \emptyset^{\bot\bot}$. Here $\top$ is the maximal element of the lattice $M$, 0 is its minimal element, $e$ is the monoidal unit, $I$ is the neutral element of the multiplicative conjunction (see hereinafter).

It is easy to get the next properties: $X^{\bot}X\subset \bot;\; X\subset X^{\bot\bot};\; X^{\bot\bot\bot} = X;\; X\multimap Y^{\bot} = (X\cdot Y)^{\bot};\; (X\cup Y)^{\bot} = X^{\bot}\cap Y^{\bot}$. From here we get that only facts may be the values and the consequents of the implication.

At facts the lattice operations of the additive conjunction $\&$ and the additive disjunction + are defined in the next way: $X \& Y = X\cap Y = (X^{\bot}\cup Y^{\bot})^{\bot};\; X + Y = (X^{\bot} \& Y^{\bot})^{\bot} = (X^{\bot}\cap Y^{\bot})^{\bot} = (X \cup Y)^{\bot\bot}$. The duality of the operations understands here as in the set theory: $\cup^{\bot} = \cap;\; \cap^{\bot} = \cup$ in which the duality means the negation.  Additive operations are interpreted as excluding option: $\oplus$ in the antecedent, and $\&$ in the consequent \cite{girar}.

At facts, multiplicative operations are also defined. They are the multiplicative conjunction $\times$ and the multiplicative disjunction $\parr$: $X\times Y = (X\cdot Y)^{\bot\bot} = (X\multimap Y^{\bot})^{\bot} = (X^{\bot}\parr Y^{\bot})^{\bot};\; X\parr Y = (X^{\bot}\cdot Y^{\bot})^{\bot} =  X^{\bot}\multimap Y$. The neutral element of the operation $\&$ is $\top$, the dual to it (neutral element of the operation +) is 0. The neutral element of the operation $\parr$ is $\bot$, the dual to it, the neutral element of the operation $\times$, is $I$.

The set of facts is divided into two classes dual to each other: the class of \textbf{open} facts $\textbf{Op}$ and the class of \textbf{closed} facts $\textbf{Cl}$. The set $\textbf{Op}$ is closed by operations + and $\times$. Its maximal element by inclusion is $I$ and the minimal one is 0. The set $\textbf{Cl}$ is closed correspondingly by operations $\&$ and $\parr$, and its maximal element is $\top$ and the minimal one is $\bot$.

Linear logic defines two \emph{exponential modalities} ! and ?. The most simple way to define the exponentials is to do it in a phase space extension called \emph{topolinear space}. The topolinear space is a phase space with a set of \emph{closed} facts $\mathscr{F}$ such that:
\begin{itemize}
\item $\mathscr{F}$ is closed under additive conjunction $\&$
\item $\mathscr{F}$ is finitely closed under multiplicative disjunction $\parr$
\item $\bot$ is the least fact (by inclusion) in $\mathscr{F}$
\item for all $X\in\mathscr{F},\;X\parr X = X.$
\end{itemize}

The set of facts $\mathscr{G}$, that is dual to $\mathscr{F}$, is referred to as a set of \emph{open} facts. They have analogous properties, with the replacements $\parr\;\rightarrow \otimes$ and $\&\;\rightarrow \oplus;\;\bot\;\rightarrow \textbf{I}$ where $I$ is the greatest open fact. The least open fact is $\textbf{0}$. The greatest closed fact, correspondingly, is $\top$. Exponentials in this space are defined as follows:
\begin{itemize}
\item $!X$ is the greatest open fact included in $X$
\item $?X$ is the least closed fact containing $X$.
\end{itemize}

As is easy to see, we have two semi-lattices: lower $\mathscr{F}$ and upper $\mathscr{G}$ such that:
\begin{itemize}
\item $!\mathscr{G} = \mathscr{G};\; ?\mathscr{F} = \mathscr{F}$
\item $(!X)^{\bot} = ?X^{\bot};\; (?X)^{\bot} = !X^{\bot}$
\end{itemize}

The propositions interpreted by open facts are said to have \emph{positive polarity} and the propositions interpreted by closed facts are said to have \emph{negative polarity}.

Those facts are valid that contain a monoidal unit. Thus we obtain:
\begin{itemize}
\item $\textbf{I}\subset \;?\textbf{I}\subset ... \subset \top$ are valid
\item and their duals $\bot \supset \;!\bot \supset ... \textbf{0}$ are false.
\end{itemize}
Therefore, in reality, we have two sets of true and false values though traditionally in the linear sequent calculus, linear logic is supposed to be 2-valued: we only can or can not infer a conclusion.

\subsection{Principles of the Linear Logic Structure Determination}\label{pr}
As it is seen, the linear implication is defined by monoidal multiplication and the choice of the
element $\bot$ which defines duality. Therefore, in order to use linear logic to choose a variant of changes in the system configuration we have to establish principles for defining multiplication and element $\bot$.

Firstly, the multiplication should be commutative, since permutations of tasks are fulfilled in parallel, should change nothing. Then, at the lattice, the element $\bot$ cannot be completely arbitrary because the lattice structure must admit mutually dual sets \emph{Cl} and \emph{Op}. Out of all possible choices we propose to choose such a variant for which the number of non-facts, i.e., elements that do not coincide with their double negations would be minimal. We do so since the consequent of linear implication, i.e., the resulting task to which the agent switches, can be a fact only. A non-fact is not a task that the system is able to fulfill now. These states were interpreted in \cite{maximov2} as the system parasite states from which it still can pass to tasks solving. These extra system states (elements of the task lattice) were corresponded to no task. If the system has such states, then it seems reasonable to require that the total number of
such possible system states is minimal. There are no such parasite states in the UAV example of Sec. \ref{3}. Hence, we will consider such UAV group tasks as non-facts, which do not fulfill at the current time, but which fulfilling is possible in future. Thereby, we demand the minimal amount of such pending tasks, i.e., that the system would be maximally active. Although, such a task cannot be an implication consequent, it may be included in it through the multiplicative and additive conjunctions in the case of the expressions are facts.

Together with the definition of $\bot$ we should also define its
dual element $I$ and mutually dual sets of open and closed facts, taking into account that $\&^{\bot} = +$. In what follows, when choosing the variants of defining multiplication, one should always take
into account that the set of open facts \emph{Op} is closed with respect to the $\otimes$ operation.

In choosing
elements dual to non-facts one should use the property

\begin{enumerate}
\item $X\subset X^{\bot\bot}$
\end{enumerate}
To define multiplication, one can use properties shown in Sec.~\ref{2}: they will imply constraints
on possible variants of multiplication determination. Thus, the following conditions are also
required:

\begin{enumerate}
\addtocounter{enumi}{1}
\item $X\multimap X \geqslant I$	
\item $I\multimap X = X$	
\item $X\multimap (B\&C) = (X\multimap B) \& (X\multimap C)$	
\item $X\multimap B^{\bot} = X\multimap C^{\bot}$ если $B^{\bot} = C^{\bot}$.
\end{enumerate}
The latter condition means that the same fact can be dual for both a fact and a non-fact.

\begin{enumerate}
\addtocounter{enumi}{5}
\item $!X\multimap \textbf{0} = \neg !X$ \cite{lafont}.
\end{enumerate}
Here $\neg !X$ is the lattice negation, i.e., the maximal lattice element which does not intersect with $!X$. Thereby, we improve  in this paper the similar heuristic condition of \cite{maximov2}.

\section{The Algorithm of the Behavior Variant Choosing}\label{pr_1}
Summing up the previous propositions, we obtain the next algorithm choosing the reconfiguration variant of an agent group:
\begin{itemize}
\item[--] Firstly, the task lattice should be built. For that purpose, every distinct task is represented by its set of elementary actions which have been necessary made to implement the task (e.g., ``move'', ``video recording'', ``signal retranslation'' and so on). The lattice is built with the sets as generators, in that case, i.e., the meets and joins are determined for all lattice elements.
\item[--] Secondly, it is necessary to define a linear logic structure at the lattice by the use of Sec. \ref{pr} principles. To define the structure, it is necessary to determine elements  $\bot$ and $I$ and, also, mutually-dual classes of open and closed facts. One can do it arbitrarily, and different choice variants determine different behavior ones. Hence, it is suggested in Sec. \ref{pr} to define the structure in such a way that the amount of non-facts would be minimal in order to reduce arbitrariness (non-facts may not be the linear consequent).
\item[--] Then, one should pick up lattice element products based on the selected structure of open and closed facts in such a way, the operation properties 1)--6) of Sec. \ref{pr} were carried out. Then, the existence of linear implication is guaranteed and its estimation we use to choose the system reconfiguration variant.  The element products are determined from the compatibility conditions of linear equations erasing from 1)--6) (see Appendix). The equation system is not wholly defined since the conditions are not enough. Hence, the multiplications are defined ambiguously.
\item[--] Finally, the tensor product of lattice elements determines the initial system state (e.g., the expression  $a\otimes (b\cup c)\otimes f$ means that one agent fulfills the task $a$, another has two intentions $b$, $c$, and the third has one intention $f$, which it fulfills). If a new task (e.g, $e$) arises, then all the implications assesses, that corresponds to switching from the initial state to all others which include the task $e$ fulfilling. E.g., the expression  $a\otimes (b\cup c)\otimes f\multimap a\otimes (b\cup e)\otimes f = d$ means that the second agent changes his intentions to $b\cup e$ and the transition value to the behavior variant is $d$.
\item[--] After that, all obtained values are compared and the maximal one is chosen. Since the element multiplications are determined ambiguously, and the implication values are not always comparable, then there is a variety in the choice of option. It is possible to use a heuristic for the final choice as in \cite{maximov2} and  to compare agent intention values as in \cite{leg}. Finally, one can use manual control as in UAV mixed group in Sec. \ref{3}.
\end{itemize}

\section{An Executor Choice in an UAV Mixed Group.}\label{3}
In the section, we consider an application method of an UAV mixed group when extinguishing forest fires. An important task in the situation is the reduction of time intervals from the very beginning to the end. One of the ways to get the solution of the problem is UAVs application \cite{gonch1}. Their main purpose is constant reconnaissance of the emergency zone. The UAV use is only beginning in current time when extinguishing forest fires, and there is even no a methodology of their using. Especially there are no precedents of UAV group use in such a situation. Let us consider a possible case of such using in future, and let along with the manned vehicle, there are object groups of different types: reconnaissance with different tasks, fire and communication brigades. However, the method is also valid for a UAV group with a ground control center.

Let us consider functioning of such a mixed group which consists of a manned vehicle and two unmanned ones. Let us suppose that one unmanned vehicle carries out reconnaissance of the fire zone $x_{2}$ in current time (the task $x_{2}$) and reconnaissance of the zone $e$ in future (the task $e$). Thus its intentions in the full task lattice are $C_{2e}$ (Fig. \ref{fig1}).

\begin{center}
\xymatrix{
  & & & \boldsymbol{\top}  \ar@{-}[dl] \ar@{-}[dll] \textbf{\ar@{-}[d]} \textbf{\ar@{-}[dr]} & & \emph{Closed} \\
   & U_{12e} \ar@{-}[dl] \ar@{-}[d] \ar@{-}[dr] & U_{13e} \ar@{-}[dll] \ar@{-}[dr] \ar@{-}[drr]& \textbf{U}_{123} \ar@{-}[dll] \textbf{\ar@{-}[dr]} \textbf{\ar@{-}[drr]}& \textbf{U}_{23e} \ar@{-}[dll] \ar@{-}[dl] \textbf{\ar@{-}[dr]} &   \\
  C_{1e} \ar@{-}[dr] \ar@{-}[drr] & \textbf{U}_{12} = \textbf{I} \textbf{\ar@{-}[d]} \textbf{\ar@{-}[drrr]} & C_{2e} \ar@{-}[d] \ar@{-}[drr] & C_{3e} \ar@{-}[dl] \ar@{-}[drr] &\textbf{U}_{13} \ar@{-}[dlll] \textbf{\ar@{-}[dr]} & \textbf{U}_{23} \textbf{\ar@{-}[d]} \ar@{-}[dl] \\
    & \textbf{U}_{1d} \textbf{\ar@{-}[d]} \textbf{\ar@{-}[drrr]} & e \ar@{-}[drr] &   & \textbf{x}_{2} \textbf{\ar@{-}[d]}  & \textbf{x}_{3} = \boldsymbol{\bot} \ar@{-}[dl] \\
   \emph{Open} & \textbf{x}_{1} \textbf{\ar@{-}[dr]}  &  &   & \textbf{d} \textbf{\ar@{-}[dll]}  &   \\
   &  & \textbf{0} &  &  &    }
   Рис. 1 Решетка задач группы БПЛА \label{fig1}
\end{center}
The second AV communicates (the task $x_{3}$), but it may engage in exploration. Thus, the system is in the state of parallel fulfilling of two tasks $x_{3}\otimes C_{2e}$ in current time. This means, in the linear logic language, that two processes are fulfilled in parallel. Let there be a need to explore the zone $x_{1}$. Therefore, the need of the system reconfiguring arises. Thus, it is necessary to find out to which state the system should pass.

The task $U_{1d}$ in the lattice of Fig. \ref{fig1} is the task of reconnaissance zone $x_{1}$, and it consists of two subtasks: subtask of aerial photography of the zone $x_{1}$ and the move subtask $d$. The latter one is included in tasks $e $, $x_{2}$ and $x_{3}$, which  also may be represented as two subtasks joins. However, for simplicity, it is not made.

Thereby, we are interested in the next system reconfiguring variants:
\begin{enumlist}[v)]
\item
$x_{3}\otimes C_{2e}\multimap x_{3}\otimes U_{12}$;

\item
$x_{3}\otimes C_{2e}\multimap x_{3}\otimes C_{1e}$;

\item
$x_{3}\otimes C_{2e}\multimap x_{1}\otimes C_{2e}$;

\item
$x_{3}\otimes C_{2e}\multimap x_{1}\otimes U_{23}$;

\item
$x_{3}\otimes C_{2e}\multimap x_{1}\otimes C_{3e}$;

\item
$x_{3}\otimes C_{2e}\multimap U_{13}\otimes C_{2e}$;

\item
$x_{3}\otimes C_{2e}\multimap U_{12}\otimes C_{3e}$;

\item
$x_{3}\otimes C_{2e}\multimap U_{23}\otimes C_{1e}$;

\item
$x_{3}\otimes C_{2e}\multimap x_{3}\otimes U_{12e}$;

\item
$x_{3}\otimes C_{2e}\multimap x_{1}\otimes U_{23e}$.
\end{enumlist}

All possible transition options from initial state $x_{3}\otimes C_{2e}$ are listed here except those (obviously pointless) in which initially delayed task begins active.

Let us define the linear logic structure at the lattice of Fig. \ref{fig1} to calculate the implications of 1v--10v. Let us choose the element $x_{3}$ as $\bot$ and the element $U_{12}$ as $I$ which is dual to $\bot$, in accordance with the algorithm of Sec. \ref{pr_1}. Correspondingly, mutually dual classes of closed and open facts are highlighted in bold font and bold lines in Fig. \ref{fig1}:

$U_{12} = I = x^{\bot}_{3} = {\bot}^{\bot}$;

$U_{1d} = U^{\bot}_{23};\;x_{1} = U^{\bot}_{23e}$;

$x_{2} = U^{\bot}_{13};\;d = U^{\bot}_{123};\;0 = {\top}^{\bot}$.

Let us choose dual for non-facts following that the property $X\subset X^{\bot\bot}$ is to be done (here $X^{\bot\bot}$ should be a fact):
\begin{gather}\label{0}
U^{\bot}_{12e} = U^{\bot}_{13e} = C^{\bot}_{1e} = 0
\end{gather}
\begin{gather}\label{x1}
C^{\bot}_{2e} = C^{\bot}_{3e} = e^{\bot} = x_{1}
\end{gather}

Using these expressions, let us rewrite 1v--10v considering that the implication consequents should be facts. Therefore, e.g.:
\begin{multline}
x_{3}\otimes C_{2e}\multimap x_{3}\otimes U_{12} = x_{3}\otimes C_{2e}\multimap (x_{3}\otimes U_{12})^{\bot\bot} = \\ =(x_{3}\otimes C_{2e}\otimes(x_{3}\otimes U_{12})^{\bot})^{\bot} = ((x_{3}\cdot C_{2e})^{\bot\bot}\cdot(x_{3}\cdot U_{12})^{\bot})^{\bot}.
\end{multline}
Similarly, we obtain:
\begin{gather}
x_{3}\otimes C_{2e}\multimap x_{3}\otimes C_{1e} = ((x_{3}\cdot C_{2e})^{\bot\bot}\cdot (x_{3}\cdot C_{1e})^{\bot})^{\bot};
\end{gather}
\begin{gather}
x_{3}\otimes C_{2e}\multimap x_{1}\otimes C_{2e} = ((x_{3}\cdot C_{2e})^{\bot\bot}\cdot (x_{1}\cdot C_{2e})^{\bot})^{\bot};
\end{gather}
\begin{gather}
x_{3}\otimes C_{2e}\multimap x_{1}\otimes U_{23} = ((x_{3}\cdot C_{2e})^{\bot\bot}\cdotp (x_{1}\cdot U_{23})^{\bot})^{\bot};
\end{gather}
\begin{gather}
x_{3}\otimes C_{2e}\multimap x_{1}\otimes C_{3e} = ((x_{3}\cdot C_{2e})^{\bot\bot}\cdot (x_{1}\cdot C_{3e})^{\bot})^{\bot};
\end{gather}
\begin{gather}
x_{3}\otimes C_{2e}\multimap U_{13}\otimes C_{2e} = ((x_{3}\cdot C_{2e})^{\bot\bot}\cdot (U_{13}\cdot C_{2e})^{\bot})^{\bot};
\end{gather}
\begin{gather}
x_{3}\otimes C_{2e}\multimap U_{12}\otimes C_{3e} = ((x_{3}\cdot C_{2e})^{\bot\bot}\cdot (U_{12}\cdot C_{3e})^{\bot})^{\bot};
\end{gather}
\begin{gather}
x_{3}\otimes C_{2e}\multimap U_{23}\otimes C_{1e} = ((x_{3}\cdot C_{2e})^{\bot\bot}\cdot (U_{23}\cdot C_{1e})^{\bot})^{\bot};
\end{gather}
\begin{gather}
x_{3}\otimes C_{2e}\multimap x_{3}\otimes U_{12e} = ((x_{3}\cdot C_{2e})^{\bot\bot}\cdot (x_{3}\cdot U_{12e})^{\bot})^{\bot};
\end{gather}
\begin{gather}
x_{3}\otimes C_{2e}\multimap x_{1}\otimes U_{23e} = ((x_{3}\cdot C_{2e})^{\bot\bot}\cdot (x_{1}\cdot U_{23e})^{\bot})^{\bot}.
\end{gather}
Thus, one can see that it is necessary to find all paired products of $x_{1}, x_{2}$ and $x_{3}$ to calculate the implications. They are given in the Appendix, and here we present only the resulting implications:
\begin{enumlist}[v)]
\item
$x_{3}\otimes C_{2e}\multimap x_{3}\otimes U_{12} = x_{1}$;

\item
$x_{3}\otimes C_{2e}\multimap x_{3}\otimes C_{1e} = \top$;

\item
$x_{3}\otimes C_{2e}\multimap x_{1}\otimes C_{2e} = x_{1}$;

\item
$x_{3}\otimes C_{2e}\multimap x_{1}\otimes U_{23} = x_{1}$;

\item
$x_{3}\otimes C_{2e}\multimap x_{1}\otimes C_{3e} = x_{1}$;

\item
$x_{3}\otimes C_{2e}\multimap U_{13}\otimes C_{2e} = \top$;

\item
$x_{3}\otimes C_{2e}\multimap U_{12}\otimes C_{3e} = \top$;

\item
$x_{3}\otimes C_{2e}\multimap U_{23}\otimes C_{1e} = \top$;

\item
$x_{3}\otimes C_{2e}\multimap x_{3}\otimes U_{12e} = \top$;

\item
$x_{3}\otimes C_{2e}\multimap x_{1}\otimes U_{23e} = x_{1}$.
\end{enumlist}
Thus, we obtain an interesting result: all variants, in which the new task completely takes the entire resource of one of the previous processes, have less value ($x_{1}$) compared to others options of the value $\top$. There are variants 3v--5v and 10v in which agents fulfilled the process turn on the new task fulfilling only\footnote{The variant 1v is technically included in this group since the process $U_{12}$ is the tensor production unit. Therefore, the state $x_{3}\otimes U_{12}$ is formally equivalent to the unique process $x_{3}$ fulfilling, i.e., the resource of one of the previous processes is again lost}. Hence, the pilot, steering group, should only to determine what is more important: constant communication support $x_{3}$ (in the case the variant 9v is preferable to 2v since it has more intentions) or urgent exploration or exploration of all zones (variants 6v--8v). Then, it is necessary to decide which tasks need communication to choose from variants 6v--8v.

It is possible to choose between the rest variants proceeding again only from the task set structure. However, we need more agent desires in the case, and not all of them should be feasible. Then, one can use the method suggested in \cite{leg}. However, the complexity of computations is greatly increased when the desires amount increases. The latter fact implies a need for a machine calculation.

\section{Conclusion}\label{concl}
Thus, the way to choose a new agent group task executor is considered. The choice is determined by the structure of the agent group task set as an emergent system property without any agent, environment and strategies modeling. Otherwise, such a choice is a real system property.

The system control should be further built following the choice taking into account aspects of specific conditions. Also, it should not contradict the purpose of the system, which the task set structure determines.

We used the lattice structure in the set of system tasks and an additional lattice elements multiplication. All this determined the linear logic structure at the lattice. Parallel fulfilling of system processes was considered as the tensor product of correspondent task lattice elements. Also, transition from executing one set of processes to another was considered as linear implication. We choose such a transition in which the implication had the greatest value.

We managed to get facilitating the decision by the pilot of mixed UAV group in the model example when extinguishing forest fires.

\section*{Appendix}  
Recall, that our multiplication is commutative. We get from the property $X\multimap X\geqslant I$ on facts:
\begin{gather}
x_{1}\multimap x_{1} = x_{1}\multimap U_{23e}^{\bot} = (x_{1}\cdot U_{23e})^{\bot} = (x_{1}x_{2} + x_{1}x_{3} + x_{1}e)^{\bot}\geqslant I = x_{3}^{\bot}
\end{gather}
Similarly:
\begin{gather}
x_{2}\multimap x_{2} = x_{2}\multimap U_{13}^{\bot} = (x_{2}x_{1} + x_{2}x_{3})^{\bot}\geqslant x_{3}^{\bot}
\end{gather}
\begin{gather}
x_{3}\multimap x_{3} = x_{3}\multimap U_{12}^{\bot} = (x_{3}I)^{\bot} = x_{3}^{\bot} = (x_{3}x_{1} + x_{2}x_{3})^{\bot}
\end{gather}
Then, we get:
\begin{gather}\label{p4}
\left\{\begin{aligned}x_{1}x_{2} + x_{1}x_{3} + x_{1}e\leqslant x_{3}\\
x_{2}x_{1} + x_{2}x_{3}\leqslant x_{3}\\
x_{3}x_{1} + x_{2}x_{3} = x_{3}\end{aligned}
\right.
\end{gather}
Similarly:
\begin{gather}\label{p5}
d\multimap d = d\multimap U_{123}^{\bot}\geqslant x_{3}^{\bot}
\end{gather}
\begin{gather}\label{p6}
U_{23}\multimap U_{23} = U_{23}\multimap U_{1d}^{\bot}\geqslant x_{3}^{\bot}
\end{gather}
Therefore:
\begin{gather}\label{p7}
\left\{\begin{aligned}dx_{1} + dx_{2} + dx_{3}\leqslant x_{3}\\
x_{1}x_{2} + x_{3}x_{1} + x_{2}d + x_{3}d\leqslant x_{3}
\end{aligned}
\right.
\end{gather}
The other variants of the property give expressions (\ref{p4}) and (\ref{p7}) by virtue of commutativity.

We obtain from the property $I\multimap X = X$:
\begin{gather}
I\multimap x_{1} = x_{1} \Rightarrow U_{12}\multimap U_{23e}^{\bot} = (U_{12}\cdot U_{23e})^{\bot} = U_{23e}^{\bot}
\end{gather}
\begin{gather}
I\multimap x_{2} = x_{2} \Rightarrow (U_{12}\cdot U_{13})^{\bot} = U_{13}^{\bot}
\end{gather}
\begin{gather}
I\multimap x_{3} = x_{3} \Rightarrow (U_{12}\cdot U_{12})^{\bot} = U_{12}^{\bot}
\end{gather}
\begin{gather}
I\multimap d = d \Rightarrow (U_{12}\cdot U_{123})^{\bot} = U_{123}^{\bot}
\end{gather}
Then:
\begin{gather}\label{p12}
\left\{\begin{aligned}x_{1}x_{2} + x_{2}x_{2} + x_{1}x_{3} + x_{2}x_{3} + x_{1}e +x_{2}e = x_{2} + x_{3} + e\\
x_{1}x_{1} + x_{2}x_{1} + x_{1}x_{3} + x_{2}x_{3} = x_{1} + x_{3}\\
x_{1}x_{1} + x_{2}x_{2} + x_{1}x_{2} + x_{2}x_{1} = x_{1} + x_{2}\\
x_{1}x_{1} + x_{2}x_{1} + x_{1}x_{2} + x_{2}x_{2} + x_{1}x_{3} +x_{2}x_{3} = x_{2} + x_{3} + x_{1}\end{aligned}
\right.
\end{gather}
In the same way:
\begin{gather}
I\multimap U_{1d} = U_{1d} \Rightarrow U_{12}\multimap U_{23}^{\bot} = (U_{12}\cdot U_{23})^{\bot} = U_{23}^{\bot}
\end{gather}
\begin{gather}
I\multimap U_{13} = U_{13} \Rightarrow (U_{12}\cdot x_{2})^{\bot} = x_{2}^{\bot}
\end{gather}
\begin{gather}
I\multimap U_{23} = U_{23} \Rightarrow (U_{12}\cdot U_{1d})^{\bot} = U_{1d}^{\bot}
\end{gather}
\begin{gather}
I\multimap U_{23e} = U_{23e} \Rightarrow (U_{12}\cdot x_{1})^{\bot} = x_{1}^{\bot}
\end{gather}
\begin{gather}
I\multimap U_{123} = U_{123} \Rightarrow (U_{12}\cdot d)^{\bot} = d^{\bot}
\end{gather}
Then:
\begin{gather}\label{p18}
\left\{\begin{aligned}x_{1}x_{2} + x_{1}x_{3} + x_{2}x_{2} + x_{2}x_{3} = x_{2} + x_{3} \\
x_{1}x_{2} + x_{2}x_{2}= x_{2}\\
x_{1}x_{1} + x_{1}d + x_{1}x_{2} + x_{2}d = x_{1} + d\\
x_{1}x_{1} + x_{2}x_{1} = x_{1}\\
x_{1}d + x_{2}d = d\end{aligned}
\right.
\end{gather}

We obtain from the property $!X\multimap 0 = \neg !X \Rightarrow Open\multimap 0 = \neg Open$:
\begin{multline}\label{19}
\left\{\begin{aligned}
U_{12}\multimap 0 = 0\\
U_{1d}\multimap 0 = 0\\
x_{1}\multimap 0 = U_{23e}\\
x_{2}\multimap 0 = x_{1}\\
d\multimap 0 = x_{1}\end{aligned}\right.
\end{multline}
Since, there are (\ref{0}) and (\ref{x1}):
\begin{gather}
0 = U_{12e}^{\bot} = U_{13e}^{\bot} = C_{1e}^{\bot} = \top^{\bot}
\end{gather}
\begin{gather}
x_{1} = C_{2e}^{\bot} = C_{3e}^{\bot} = e^{\bot} = U_{23e}^{\bot}
\end{gather}
Hence:
\begin{gather}
I\multimap 0 = 0 = U_{12}\multimap U_{12e}^{\bot} = (U_{12}\cdot U_{12e})^{\bot} = (I\cdot U_{12e})^{\bot} = U_{12e}^{\bot}
\end{gather}
Similarly:
\begin{gather}
(U_{12}\cdot U_{13e})^{\bot} = U_{13e}^{\bot}
\end{gather}
\begin{gather}
(U_{12}\cdot C_{1e})^{\bot} = C_{1e}^{\bot}
\end{gather}
Hence, we obtain:
\begin{multline}\label{p25}
\left\{\begin{aligned}
x_{1}x_{1} + x_{1}x_{2} + x_{1}e + x_{2}x_{1} + x_{2}x_{2} + x_{2}e = x_{1} + x_{2} + e\\
x_{1}x_{1} + x_{1}x_{3} + x_{1}e + x_{2}x_{1} + x_{2}x_{3} + x_{2}e = x_{1} + x_{3} + e\\
x_{1}x_{1} + x_{1}e + x_{2}x_{1} + x_{2}e = x_{1} + e
\end{aligned}\right.
\end{multline}
For the other expressions (\ref{19}), we obtain  in the same way (the sign $\vee$ divides possible variants, here and below):
\begin{multline}\label{p26}
\left\{\begin{aligned}
x_{1}x_{1} + x_{1}x_{2} + x_{1}e + dx_{1} + dx_{2} + de = x_{1} + x_{2} + e \vee x_{1} + x_{3} + e \vee x_{1} + e \vee x_{1} + x_{2} + x_{3} + e \\
x_{1}x_{1} + x_{1}x_{3} + x_{1}e + dx_{1} + dx_{3} + de = x_{1} + x_{2} + e \vee x_{1} + x_{3} + e \vee x_{1} + e \vee x_{1} + x_{2} + x_{3} + e\\
x_{1}x_{1} + x_{1}e + dx_{1} + de = x_{1} + x_{2} + e \vee x_{1} + x_{3} + e \vee x_{1} + e \vee x_{1} + x_{2} + x_{3} + e
\end{aligned}\right.
\end{multline}
\begin{multline}\label{p27}
\left\{\begin{aligned}
x_{1}x_{1} + x_{1}x_{2} + x_{1}e = x_{1} \\
x_{1}x_{1} + x_{1}x_{3} + x_{1}e  = x_{1}\\
x_{1}x_{1} + x_{1}e = x_{1}
\end{aligned}\right.
\end{multline}
\begin{multline}\label{p28}
\left\{\begin{aligned}
x_{2}x_{1} + x_{2}x_{2} + x_{2}e = x_{3} + x_{2} + e \vee x_{2} + e \vee x_{3} + e \vee e\\
x_{2}x_{1} + x_{2}x_{3} + x_{2}e  = x_{3} + x_{2} + e \vee x_{2} + e \vee x_{3} + e \vee e\\
x_{2}x_{1} + x_{2}e = x_{3} + x_{2} + e \vee x_{2} + e \vee x_{3} + e \vee e
\end{aligned}\right.
\end{multline}
\begin{multline}\label{p29}
\left\{\begin{aligned}
dx_{1} + dx_{2} + de = x_{3} + x_{2} + e \vee x_{2} + e \vee x_{3} + e \vee e\\
dx_{1} + dx_{3} + de  = x_{3} + x_{2} + e \vee x_{2} + e \vee x_{3} + e \vee e\\
dx_{1} + de = x_{3} + x_{2} + e \vee x_{2} + e \vee x_{3} + e \vee e
\end{aligned}\right.
\end{multline}
The case $X\multimap {\top}^{\bot}$ is obtained from the others.

Finally, from property 5 (Sec. \ref{pr}), the next expression should be held for arbitrary $X$:
\begin{gather}
X\multimap C_{2e}^{\bot} = X\multimap C_{3e}^{\bot} = X\multimap e^{\bot} = X\multimap U_{23e}^{\bot} = X\multimap x_{1}
\end{gather}
This means, that:
\begin{gather}\label{p31}
(X\cdot C_{2e})^{\bot} = (X\cdot C_{3e})^{\bot} = (X\cdot e)^{\bot} = (X\cdot U_{23e})^{\bot}
\end{gather}

From the requirement of compatibility of expressions (\ref{p4}), (\ref{p7}), (\ref{p12}), (\ref{p18}), (\ref{p25})--(\ref{p29}) and (\ref{p31}), we get:
\begin{multline}
\left\{\begin{aligned}
x_{1}x_{1} = x_{1}\\
x_{1}x_{2} = 0\\
x_{1}x_{3} = 0\\
x_{1}d = 0 \vee d\\
x_{1}e = 0\\
x_{2}x_{2} = x_{2}\\
x_{2}x_{3} = x_{3}\\
x_{2}d = d \vee 0\\
x_{2}e = e\\
x_{1}d + x_{2}d = d\\
x_{3}x_{3} = e \vee x_{2} \vee x_{3} \vee U_{23} \vee C_{2e} \vee C_{3e} \vee U_{23e}\\
x_{3}e = e \vee C_{2e} \vee C_{3e} \vee U_{23e}\\
x_{3}C_{3e} = e \vee C_{2e} \vee C_{3e} \vee U_{23e}\\
x_{3}d = x_{3} \vee d \vee 0\\
de = U_{23e} \vee C_{2e} \vee C_{3e} \vee e\\
ee = e \vee C_{2e} \vee C_{3e} \vee U_{23e}
\end{aligned}\right.
\end{multline}
Thus, from (3)--(12), we finally get for 1v--10v:
\begin{enumlist}[v)]
\item
$x_{3}\otimes C_{2e}\multimap x_{3}\otimes U_{12} = ((x_{3} + x_{3}e)^{\bot\bot}\cdot (x_{3})^{\bot})^{\bot} = ((C_{3e}\vee U_{23e})^{\bot\bot}\cdot U_{12})^{\bot} = U_{23e}^{\bot} = x_{1}$;

\item
$x_{3}\otimes C_{2e}\multimap x_{3}\otimes C_{1e} = (U_{23e}\cdot x_{1})^{\bot} = \top$;

\item
$x_{3}\otimes C_{2e}\multimap x_{1}\otimes C_{2e} = (U_{23e}\cdot \top)^{\bot} = (U_{23e} + C_{3e} \vee U_{23e})^{\bot} = x_{1}$;

\item
$x_{3}\otimes C_{2e}\multimap x_{1}\otimes U_{23} = (U_{23e}\cdot \top)^{\bot} = x_{1}$;

\item
$x_{3}\otimes C_{2e}\multimap x_{1}\otimes C_{3e} = (U_{23e}\cdot \top)^{\bot} = x_{1}$;

\item
$x_{3}\otimes C_{2e}\multimap U_{13}\otimes C_{2e} = (U_{23e}\cdot (C_{3e} \vee U_{23e})^{\bot})^{\bot} = \top$;

\item
$x_{3}\otimes C_{2e}\multimap U_{12}\otimes C_{3e} = (U_{23e}\cdot C_{3e}^{\bot})^{\bot} = \top$;

\item
$x_{3}\otimes C_{2e}\multimap U_{23}\otimes C_{1e} = (U_{23e}\cdot (e \vee C_{2e} \vee C_{3e} \vee U_{23e})^{\bot})^{\bot} = (U_{23e}\cdot x_{1})^{\bot} = \top$;

\item
$x_{3}\otimes C_{2e}\multimap x_{3}\otimes U_{12e} = (U_{23e}\cdot x_{1})^{\bot} = \top$;

\item
$x_{3}\otimes C_{2e}\multimap x_{1}\otimes U_{23e} = (U_{23e}\cdot \top)^{\bot} = x_{1}$.
\end{enumlist}

\end{document}